\documentclass[preprint,12pt]{elsarticle}

\usepackage{amssymb}
\usepackage{amsmath}

\usepackage{graphicx}
\usepackage{multirow}
\usepackage{caption}
\usepackage{subcaption}
\usepackage{arydshln}
\usepackage{adjustbox}
\usepackage{tcolorbox}

\usepackage[T1]{fontenc}
\usepackage[T5]{fontenc}

\usepackage{pifont}
\newcommand{\cmark}{\ding{51}}%
\newcommand{\xmark}{\ding{55}}%

\usepackage[normalem]{ulem}
\useunder{\uline}{\ul}{}

\journal{International Journal of Artificial Intelligence in Education}

\begin{document}

\begin{frontmatter}



\title{ViMultiChoice: Toward a Method That Gives Explanation for Multiple-Choice Reading Comprehension in Vietnamese} 


\author[1,3,4]{Trung Tien Cao}
\ead{22521553@gm.uit.edu.vn}

\author[1,3,4]{Lam Minh Thai}
\ead{22520745@gm.uit.edu.vn}

\author[1,3,4]{Nghia Hieu Nguyen}
\ead{nghiangh@uit.edu.vn}

\author[2,3,4]{Duc-Vu Nguyen}
\ead{vund@uit.edu.vn}

\author[1,3,4]{Ngan Luu-Thuy Nguyen}
\ead{ngannlt@uit.edu.vn}

\affiliation[1]{organization={Faculty of Information Science and Engineering}}

\affiliation[2]{organization={Faculty of Computer Science}}

\affiliation[3]{organization={University of Information Technology, Vietnam National University}}

\affiliation[4]{organization={Vietnam National University}, city={Ho Chi Minh city}, country={Viet Nam}}

\begin{abstract}
Multiple-choice Reading Comprehension (MCRC) models aim to select the correct answer from a set of candidate options for a given question. However, they typically lack the ability to explain the reasoning behind their choices. In this paper, we introduce a novel Vietnamese dataset designed to train and evaluate MCRC models with explanation generation capabilities. Furthermore, we propose ViMultiChoice, a new method specifically designed for modeling Vietnamese reading comprehension that jointly predicts the correct answer and generates a corresponding explanation. Experimental results demonstrate that ViMultiChoice outperforms existing MCRC baselines, achieving state-of-the-art (SotA) performance on both the ViMMRC 2.0 benchmark and the newly introduced dataset. Additionally, we show that jointly training option decision and explanation generation leads to significant improvements in multiple-choice accuracy. To support reproducibility, the code and dataset will be publicly released upon acceptance of this paper.
\end{abstract}




\end{frontmatter}


\section{Introduction}

Multiple-choice Reading Comprehension (MCRC) plays an important role in a wide range of applications, including virtual assistants and conversational agents. Recent advances in Information Retrieval (IR) \cite{ir-4} and Large Language Models (LLMs) \cite{qwen2} have opened up new directions for research and development in this area.

While MCRC research has been extensively studied for high-resource languages such as English and Chinese, work on low-resource languages, including Vietnamese, remains limited. Existing studies on Machine Reading Comprehension (MRC) in general, and MCRC in particular, for Vietnamese have primarily focused on introducing large-scale benchmarks across different domains, such as healthcare news \cite{vinewsqa,vicoqa} and education \cite{vimmrc}. In addition, most proposed Vietnamese MCRC models rely on complex architectures that integrate multiple components, including Natural Language Inference (NLI) \cite{nli-1,nli-2} and multi-hop question answering \cite{multi-hop-1,multi-hop-2}. These approaches are computationally expensive and often overlook the linguistic characteristics specific to Vietnamese. Moreover, MCRC methods that are capable of providing explanations for their predictions are more desirable, as they offer greater transparency and are more suitable for real-world applications.

To address these limitations, we propose a novel MCRC approach that is specifically tailored to the linguistic features of Vietnamese and is capable of generating explanations for its selected option. To evaluate the effectiveness of the proposed method, we introduce the ViRCSoSciD (\textbf{R}eading \textbf{C}omprehension for \textbf{So}cial \textbf{Sci}ence \textbf{D}isciplines in \textbf{Vi}etnamese) dataset, which consists of \textbf{12,819} multiple-choice questions. The questions are derived from four social science subjects in the Vietnamese high school curriculum. Each question contains four options and is accompanied by a human-annotated explanatory text. ViRCSoSciD serves as a high-quality benchmark for evaluating explanation-aware MCRC systems in Vietnamese, including our proposed approach as well as future models developed by other researchers.

Our contributions can be summarized as follows:
\begin{enumerate}
    \item We introduce the first Vietnamese MCRC dataset that includes human-annotated explanations for the correct option, enabling the evaluation of explanation-aware MCRC systems.
    \item We propose \textbf{ViMultiChoice}, a novel MCRC method tailored to Vietnamese linguistic characteristics, with the ability to generate explanations for its predictions.
\end{enumerate}
Through these contributions, we aim to provide a foundational framework for future research on MCRC modeling for Vietnamese and other low-resource languages.

\section{Related Works}

\subsection{Datasets}

Recent advances in Machine Reading Comprehension (MRC), particularly in the multiple-choice setting (MC-MRC), have been driven by the release of large-scale benchmark datasets designed to evaluate increasingly complex reasoning abilities of language models. Early datasets such as MCTest \cite{mctest} focused on basic reading comprehension but were limited in scale and reasoning depth, while RACE \cite{race} marked a significant step forward by introducing longer and more complex passages from high school examinations. However, subsequent studies revealed that many RACE questions could still be solved using shallow heuristics or by exploiting statistical answer-position biases. To address these limitations, newer benchmarks have been developed to target higher-level reasoning, including formal logical reasoning in ReClor \cite{reclor} or commonsense reasoning in CommonsenseQA \cite{commonsenseqa}, and multi-step scientific reasoning in OpenBookQA \cite{OpenBookQA}. More recent datasets, such as MMMU \cite{mmmu} and MathVista \cite{mathvista}, further extend MC-MRC to expert-level academic and mathematical domains, reflecting a clear shift from surface-level text understanding toward structured, multi-step reasoning.

In parallel, research on MC-MRC for Vietnamese has gained increasing attention, although the data ecosystem remains relatively limited compared to high-resource languages. Early Vietnamese MRC efforts such as UIT-ViQuAD \cite{viquad} focused on extractive question answering and did not capture the challenges of multiple-choice reasoning. Subsequent datasets, including ViMMRC 2.0 \cite{vimmrc}, addressed this gap in educational contexts by introducing literary texts with higher-level reasoning requirements and semantically similar distractors. Despite these advances, existing Vietnamese datasets still face limitations in terms of reasoning depth, domain coverage, and explanation-based evaluation, highlighting the need for more comprehensive and explanation available MC-MRC benchmarks.

\subsection{Methods}

Multiple-choice Machine Reading Comprehension (MC-MRC) poses unique challenges compared to extractive or generative question answering, as it requires models to compare semantically similar options, perform multi-step reasoning, and eliminate distractors. Early approaches to MC-MRC largely inherited traditional MRC architectures based on sequential text representations, employing RNN-based models such as LSTMs or GRUs to independently encode the context, question, and each answer option. Representative methods from this stage include Memory Network–based models \cite{mmm}, which treat textual units as memory slots and select answers by measuring relevance between question–option pairs and contextual memory, as well as CNN-based approaches \cite{cnn} that capture local semantic patterns. However, these methods generally lack explicit mechanisms for modeling fine-grained interactions among answer choices and struggle with long-range dependencies and complex reasoning.

To overcome these limitations, subsequent research shifted toward interaction- and matching-based methods that explicitly model semantic relationships between the context, question, and answer options. Co-matching frameworks such as Co-Match \cite{co-matching} and MMN \cite{mmm} introduced joint and multi-level matching mechanisms to capture both local and global semantic alignments, while hierarchical and multi-paragraph models such as HAF \cite{HierarchicalAttention} and DynSAN \cite{dynsan} addressed the challenges posed by long or multi-document contexts through hierarchical and dynamic attention mechanisms. In parallel, option-centric approaches including OCN \cite{OCN} and ElimiNet \cite{ElimiNet} focused on direct comparison and elimination of answer choices to better distinguish closely related distractors, mimicking human test-taking strategies. More recent pre-Transformer methods further enhanced matching architectures by modeling higher-order interactions, such as DCMN+ \cite{dcmn} and CTM \cite{ctm}, which introduced strengthened co-matching and triple-matching mechanisms among context, question, and options. Overall, the evolution of MC-MRC methods reflects a progression from independent representation learning toward increasingly sophisticated interaction modeling and reasoning strategies.

\section{The ViRCSoSciD Dataset}

\begin{figure*}[t]
    \centering
    \begin{subfigure}[b]{0.35\textwidth}
        \centering
        \includegraphics[width=\textwidth]{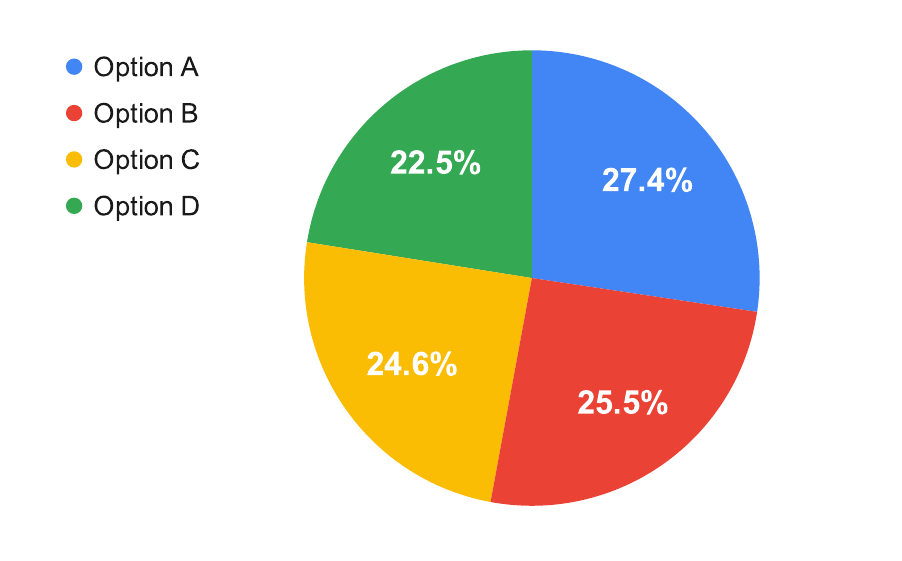}
        \caption{}
        \label{fig:data-statistics-a}
    \end{subfigure}
    \begin{subfigure}[b]{0.3\textwidth}
        \centering
        \includegraphics[width=\textwidth]{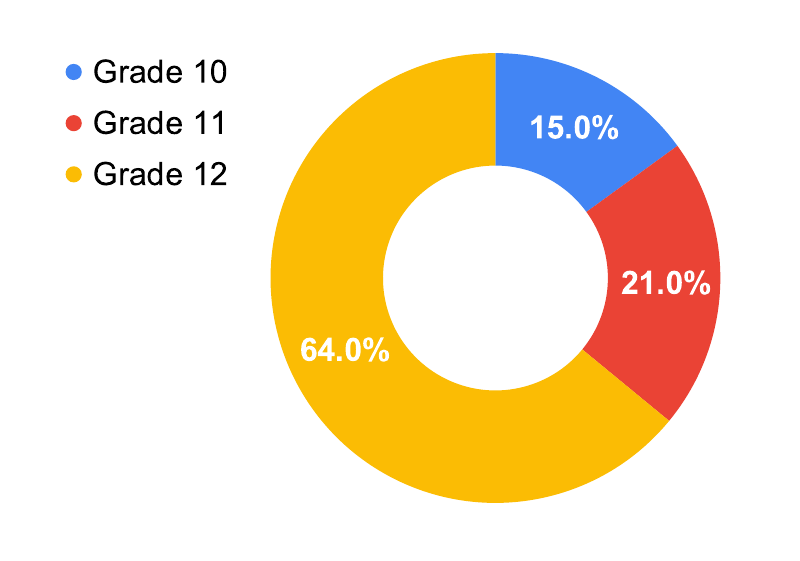}
        \caption{}
        \label{fig:data-statistics-b}
    \end{subfigure}
    \begin{subfigure}[b]{0.3\textwidth}
        \centering
        \includegraphics[width=\textwidth]{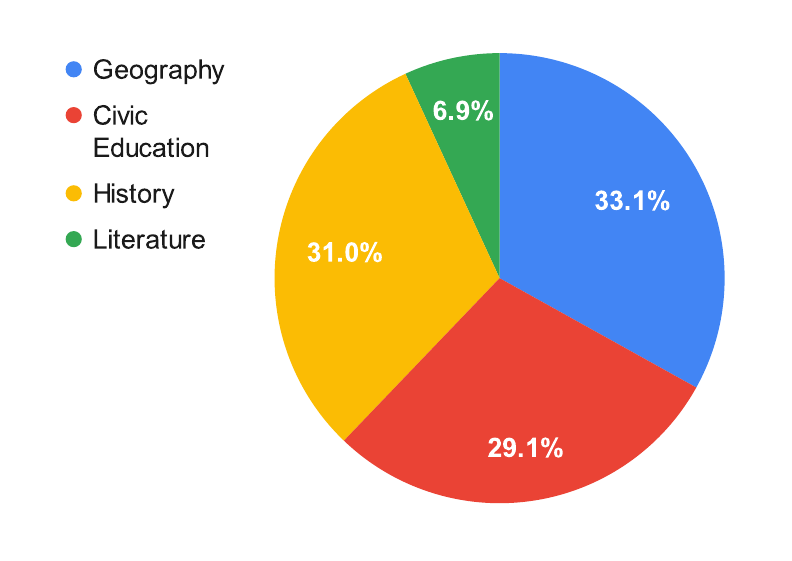}
        \caption{}
        \label{fig:data-statistics-c}
    \end{subfigure}
    \caption{Statistics of questions among options, grades, and subjects in the ViRCSoSciD dataset.
    \label{fig:data-statistics}}
\end{figure*}

\subsection{Dataset Construction}

\subsubsection{Multiple-choice Questions Collection}

The multiple-choice questions were collected from well-established educational platforms that provide extensive resources for high school students. In particular, we crawled questions from websites which offer large-scale and open educational questions covering the entire high school curriculum across all grade levels. We focus exclusively on multiple-choice questions designed for high school students (Grades \textbf{10}, \textbf{11}, and \textbf{12}) in four social science subjects: \textbf{Literature}, \textbf{History}, \textbf{Geography}, and \textbf{Civic Education}. Each question contains four answer options, following the official format specified by the Vietnamese Ministry of Education and Training for high school graduation and university entrance examinations.

\subsubsection{Answers and Explanation Annotation}

The correct answers and their corresponding explanations were annotated by teachers from Nguyen Huu Huan high school. In total, 7 high school teachers participated in the annotation process, providing answers and explanations for all questions. For each subject, between two and three teachers served as annotators. All annotators have at least 10 years of teaching experience at the high school level and a minimum of 5 years of experience in preparing students for university entrance examinations. Within each subject, annotators cross-validated one another’s annotations to assess inter-annotator consistency. At the conclusion of the annotation process, the Cohen’s kappa inter-annotator agreement \cite{cohen-kappa} reached 96\%, indicating a very high level of agreement among teachers across the four subjects in the ViRCSoSciD dataset.

\subsubsection{Textbooks Collection}

We collected official textbooks of the Vietnam Ministry of Training and Education which are publicly available in the PDF format. These books are converted into plain text format by using Mathpix Snipping\footnote{https://mathpix.com/} tool. The textbooks in new format might contain several errors relevant to the OCR recognition of Maxthpix. To this end, we manually verify every page of all textbooks to ensure the correct conversion of the tool. Moreover, to improve data quality and relevance, we remove practice-oriented lessons that contain limited knowledge. Finally, the corpus underwent a comprehensive preprocessing pipeline, including text normalization, spelling correction, and the removal of non-informative elements such as footnotes and figure descriptions.

\subsection{Overall Statistics}

\begin{figure}
    \centering
    \includegraphics[width=\linewidth]{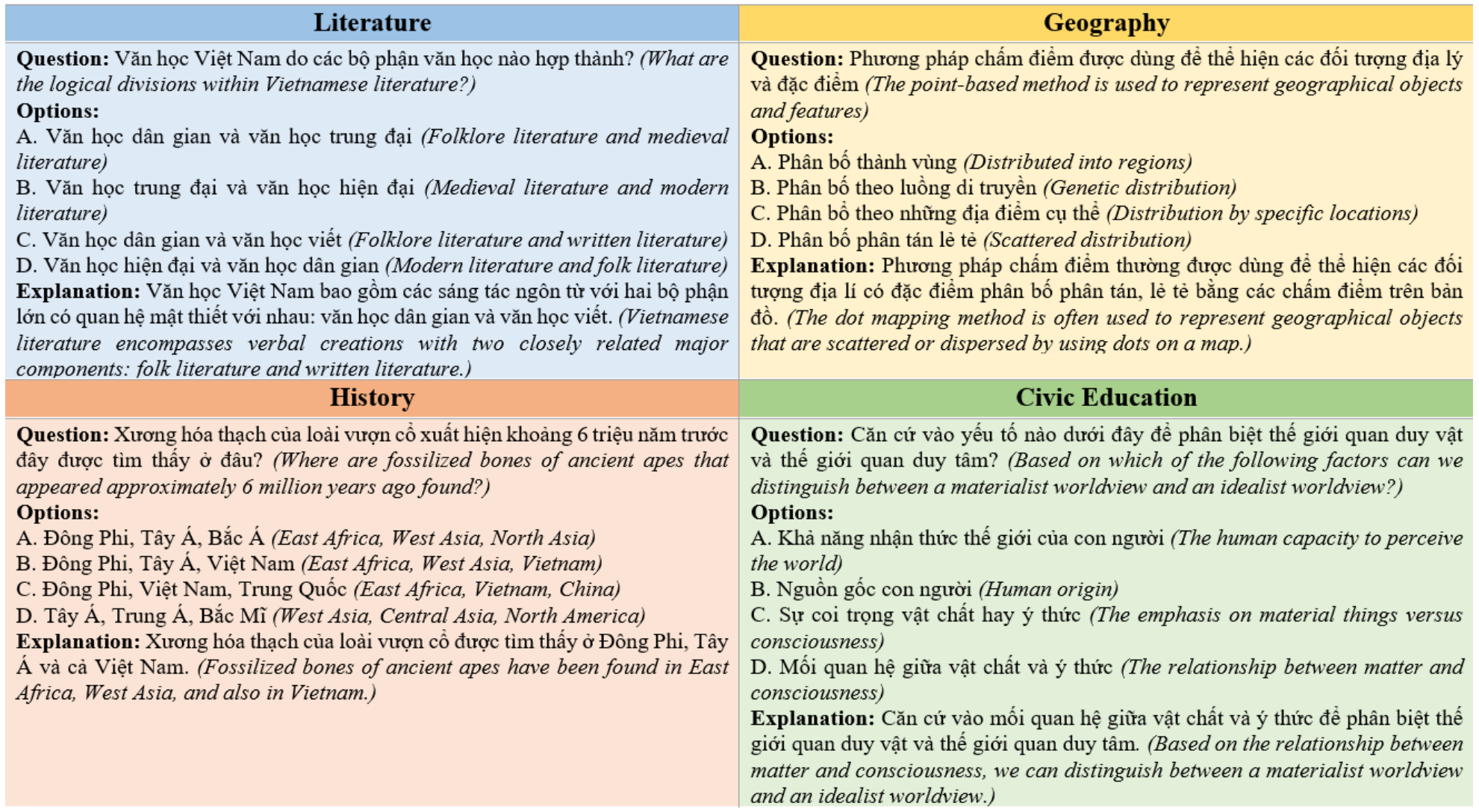}
    \caption{Some samples from the ViRCSoSciD dataset.}
    \label{fig:examples}
\end{figure}

Overall, our dataset contains \textbf{12,819} multiple-choice questions. As shown in Figure~\ref{fig:data-statistics}, approximately half of the questions are designed for Grade~12 students, while Grade~10 accounts for the smallest proportion. Questions targeting Grade~12 students are generally more complex, as they often require comprehensive knowledge accumulated across all three high school grades to identify relevant information and determine the correct answer. These questions are primarily intended to support preparation for the Vietnamese university entrance examination (Figure \ref{fig:examples}).

As a result, the questions in our dataset are more challenging than those in existing Vietnamese reading comprehension corpora, posing greater difficulty for deep learning–based MCRC models.

\subsection{On the Bias Choice Problem}

\begin{figure}
    \centering
    \includegraphics[width=\textwidth]{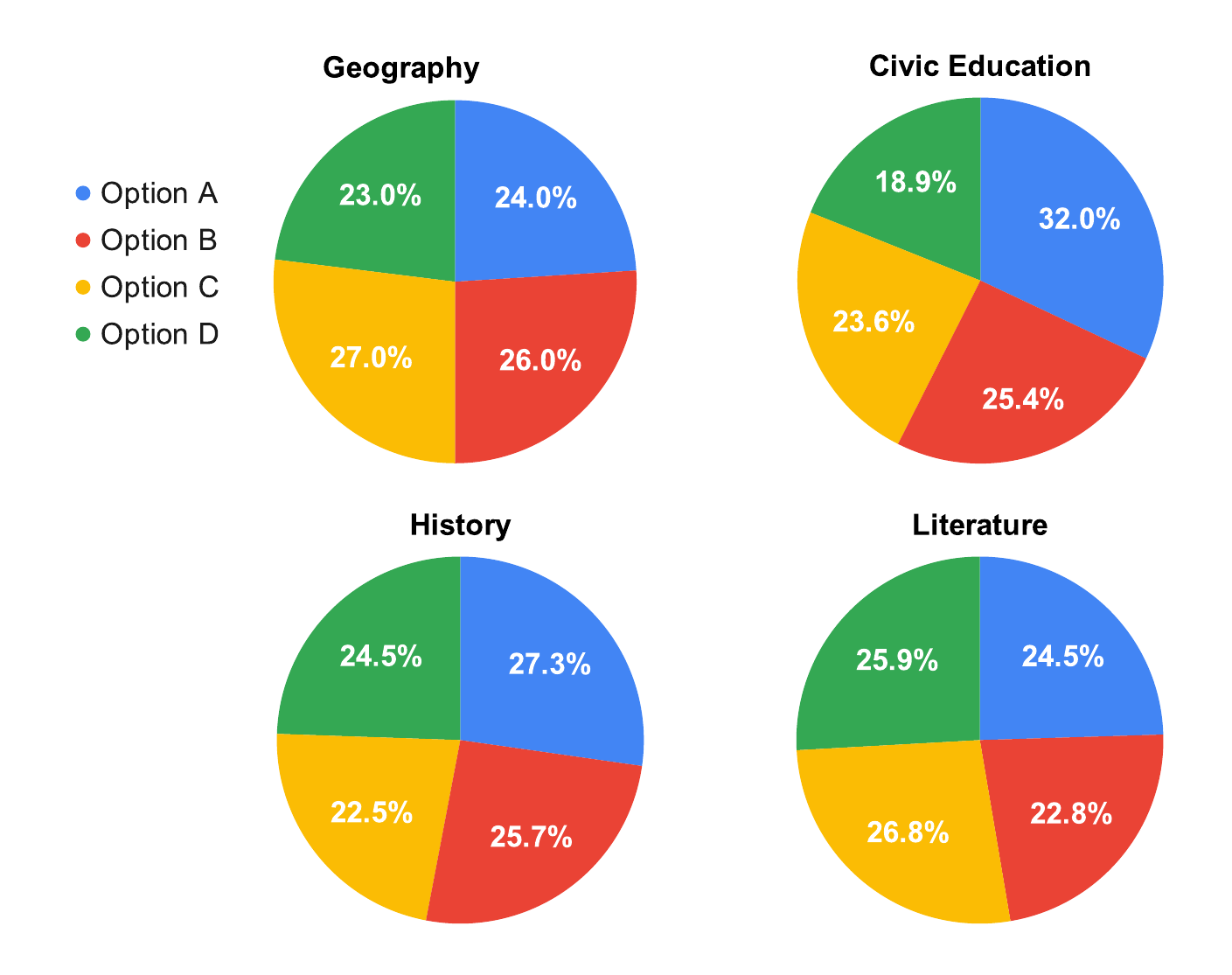}
    \caption{Distributions of options among subjects in the ViRCSoSciD dataset.}
    \label{fig:answer-subject-dis}
\end{figure}

A significant challenge in constructing multiple-choice question answering datasets is the presence of choice bias \cite{bias-choice}, where models tend to rely on the prior distribution of options (A, B, C, or D) rather than truly understanding the content of the question. This bias undermines the model's ability to reason correctly and can lead to inaccuracies in performance evaluations. 
To address this, we examined the frequency distribution of correct options and applied a de-biasing strategy by shuffling the order of choices. This process reduces the impact of label position shortcuts and creates a more balanced dataset, ensuring that model performance is based on true knowledge and reasoning rather than statistical patterns.

According to Figure \ref{fig:answer-subject-dis}, the distribution of four options is mostly uniform. There is an exception on Civic Education where the option A has the highest percentage. However, the distribution of four option is still uniform in general (Figure \ref{fig:data-statistics-a}), which directly limit the effect of the bias choice problem.

\section{Methodology}

\begin{figure*}
    \centering
    \includegraphics[width=\linewidth]{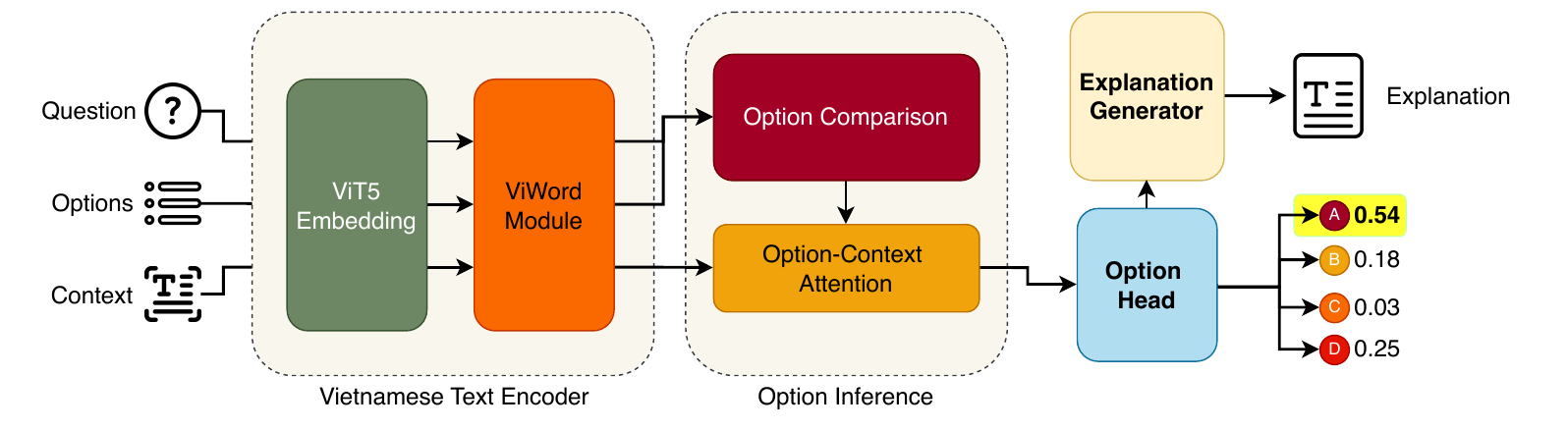}
    \caption{Architecture of the ViMultiChoice method. \textbf{MSA} stands for \textbf{Multihead Self-Attention} module.}
    \label{fig:vimultichoice}
\end{figure*}

\subsection{Overall Architecture}

Our proposed ViMultiChoice (\textbf{Multi}ple-\textbf{Choice} Reading Comprehension in \textbf{Vi}etnamese) is developed based on the OCN \cite{OCN} method with the advancement in Vietnamese language modeling and explanation generation. In particular, the ViMultiChoice has three main components: Vietnamese Text Encoder, Option Inference and Explanation Generator (Figure \ref{fig:vimultichoice}).

\subsection{Context Retrieval Module}


We applied an information retrieval pipeline to retrieve supporting context for multiple-choice questions. The process begins by filtering knowledge based on the question's subject, followed by splitting the relevant content into sentence-level units. To enhance retrieval precision, we construct queries by averaging the representations of the question and its corresponding answer options. This approach ensures that the retrieved context captures the collective semantics of the entire multiple-choice item, rather than being limited to the question's isolated information.

In this study, we examined the effectiveness of SotA text retrievers to collect relevant context from text books for the questions of the ViRCSoSciD corpus. In particular, we conducted experiments for BM25 \cite{bm25}, MiniLML12 \cite{sbert}, MultiE5 \cite{multie5}, VietBGE \cite{vietbge}, and VietBGEMultiE5 which is the combination of VietBGE and MultiE5 pretrained embeddings. Experimental results show that VietbGE is the most effective one. From that on, MCRC methods on the ViRCSoSciD were trained with context retrieved by VietBGE from the Vietnamese standard textbooks.

\subsection{Vietnamese Text Encoder}

\begin{figure}[htp]
    \centering
    \includegraphics[width=\linewidth]{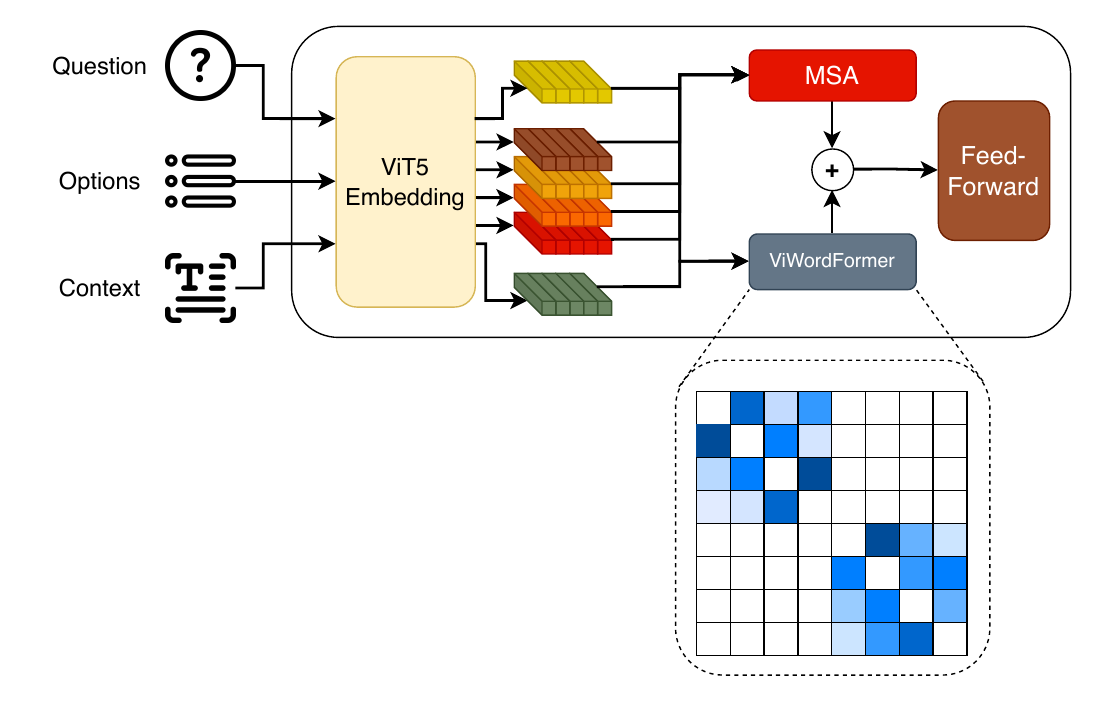}
    \caption{The architecture of Vietnamese Text Encoder. \textbf{MSA} stands for \textbf{Multihead Self-Attention} module}
    \label{fig:viwordmodule}
\end{figure}

Vietnamese Text Encoder aims to encode and represent Vietnamese texts in questions, options, and contexts for answer inference and explanation generation. This module consists of two components: (1) a pretrained language model for Vietnamese and (2) the ViWordFormer module (Figure \ref{fig:viwordmodule}).

Given a question $Q = \{w^q_1, w^q_2, ..., w^q_{n_q}\}$, four options $O = \{O_1, O_2, O_3, O_4\}$ with $O_k = \{w_1^{O_k}, w_2^{O_k}, ..., w_{n_{O_k}}^{O_k}\}$, and the retrieved context $P = \{w_1^p, w_2^p, ..., w_{n_P}^p\}$ as the input, we obtain their embedded vectors by the embedding layers $Emb$ of a pretrained language model for Vietnamese:

$$
    f_Q = Emb(Q) = [f^Q_1, f^Q_2, ..., f^Q_{n_Q}]^T \in \mathbb{R}^{n_Q \times d}
$$

$$
    f_{O_k} = Emb(O_k) = [f^{O_k}_1, f^{O_k}_2, ..., f^{O_k}_{n_Q}]^T \in \mathbb{R}^{n_{O_k} \times d}
$$

$$
    f_P = Emb(P) = [f^P_1, f^P_2, ..., f^P_{n_P}]^T \in \mathbb{R}^{n_Q \times d}
$$

We inspired by the recent study \cite{viwordformer} to have a module that describes the linguistic features of Vietnamese phrasal lexemes for better understanding and representation modeling in this language. 
In particular, Vietnamese has compound words having two syllables separated by spaces in writing. To this end, \cite{treebank} claimed that word segmentation is a linguistic feature of Vietnamese. However \cite{viwordformer} argue that such compound words are actually phrases in Vietnamese, hence word segmentation is a particular way of modeling Vietnamese phrases. Following this study, we propose the ViWordFormer module which describes phrasal lexemes in Vietnamese texts.

Given the embedded vectors of the input text $w = \{w_1, w_2, ..., w_{n_w} \}$ as $f_w = [f^w_1, f^w_2, ..., f^w_{n_w}]^T \in \mathbb{R}^{n_w \times d}$, ViWordFormer first has the multihead self-attention module of Transformer \cite{Vaswani2017AttentionIA} for determining the Attention Score matrix $A$ of every token in $f_w$:

$$
    A = softmax\left(\frac{(W_qf_w)(W_k f_w)^T}{\sqrt{d}}\right) \in \mathbb{R}^{n_w \times n_w}
$$
where $W_q \in \mathbb{R}^{d \times d}$ and $W_k \in \mathbb{R}^{d \times d}$ are the learnable parameters.

However, there is no linguistic prior constraints on the attention weights of any two tokens in $w$, which means there can be a high correlation between two tokens belonging to two different phrases. To this end, ViWordFormer introduces the Phrasal Score matrix $P$ to re-correct the connections among tokens determined by $A$:

\begin{equation} \label{eq:new_attention}
    A' = A \odot P
\end{equation}
Then the refined vector of $f_w$ is calculated as follows:

\begin{equation}
    f_w' = A' (W_v f_w)
\end{equation}
where $W_v \in \mathbb{R}^{d \times d}$ is the learnable parameter.

As mentioned above, the Phrasal Score matrix $P = (P_{ij})$ describes Vietnamese phrasal lexemes. Each $P_{ij}$ measure the probability of $w_i, w_{i+1}, ..., w_j$ are in the same phrase. This probability is determined as:

\begin{equation}
    P_{ij} = \prod_{k=i}^{j-1} P_k
\end{equation}
where $P_k$ is the probability of $w_k$ and $w_{k+1}$ are in the same phrase or they form a compound word \cite{treebank}. This term is calculated as the geometric average of two terms $pr_{k, k+1}$ and $pr_{k+1, k}$:

\begin{equation} \label{eq:p_k}
    P_k = (pr_{k, k+1} \cdot pr_{k+1, k})^\frac{1}{2}
\end{equation}

The term $pr_{k, k+1}$ in equation (\ref{eq:p_k}) describes the probability of the words $w_k$, represented by $f^T_k$, has the semantic relation with its neighbor words $w_{k-1}$ and $w_{k+1}$ represented by two embedded vectors $f^T_{k-1}$ and $f^T_{k+1}$, respectively. This probability is obtained as follows:

\begin{equation}
    pr_{k-1, k}; pr_{k, k+1} = softmax(r_{k-1, k}; r_{k, k+1})
\end{equation}
where $r_{k, k+1}$ describe the semantic relation of two words represented by $f_k^T$ and $f_{k+1}^T$:

\begin{equation}
    r_{k, k+1} = b(f_k^w, f_{k+1}^w) = (f_k^T)^T W_b f_{k+1}^T
\end{equation}
with $W_b \in \mathbb{R}^{d \times d}$ is the learnable parameters representing the bilinear function $b: \mathbb{R}^d \times \mathbb{R}^d \mapsto \mathbb{R}$.

However, as $P_k \in [0, 1]$ hence $P_{ij} \rightarrow 0$ rapidly when $|i - j| \rightarrow \infty$. To this end, we convert $P_{ij}$ from the product space to the summation space via the combination of $exp$ and $ln$ bijection:

$$
P_{ij} = exp(ln(P_{ij})) = exp\left(ln\left(\prod_{k=i}^{j-1} P_k\right)\right)
     = exp\left( \sum_{k=i}^{j-1} ln P_k \right) 
$$
Having these two terms $A$ and $P$ are all in the form of exponential functions, we describe the intuition of adding $P$ to $A$ via the element-wise operator $\odot$ as in (\ref{eq:new_attention}) rather than the addition operator.

We provide three ViWordFormer module to modeling Vietnamese phrasal structures for questions $f_Q$, options $f_{O_k}$, and context $f_P$. These features are then forwarded to the Encoder of the Pretrained Language model to achieve the representation for the input.

\subsection{Option Inference Module}

\begin{figure}[htp]
    \centering
    \includegraphics[width=0.75\linewidth]{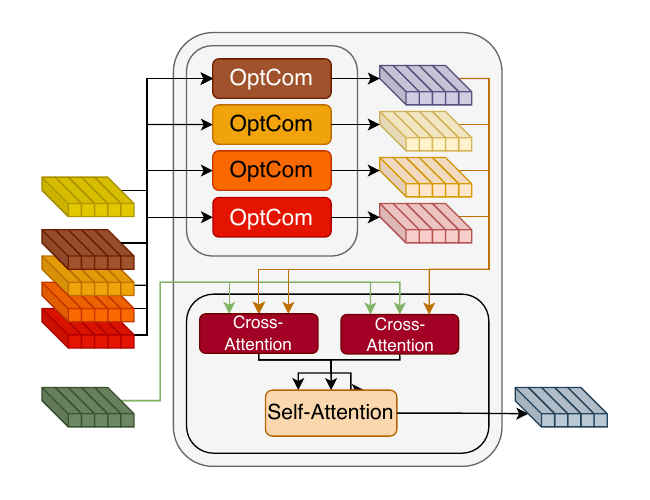}
    \caption{The Option Inference module. \textbf{OptCom} stands for \textbf{Option Comparison}.}
    \label{fig:opt-module}
\end{figure}

Given the features $f_Q, f_{O_k}, f_P$ from the Vietnamese Word Encoder, we develop the Option Inference Module which is mostly based on the OCN \cite{OCN} method for reasoning and deciding options. This module contains four \textbf{Option Comparison module}. Each of them determines the feature vector $f_k$ representing for each option $O_k$ after considering the information from other options and the context (Figure \ref{fig:opt-module}).

As the question and the options share the many relevant information for the final decision, we concatenate the question into every option:

$$
    f_{O_k}^Q = [f_Q; f_{O_k}] \in \mathbb{R}^{n_{Qk} \times d}, k \in \{1, 2, 3, 4\}
$$
where $[;]$ denotes the concatenation operator along the last dimension and $n_{Qk} = n_Q + n_k$. First, we compare any option $O_k$ with another option $O_l (l \ne k)$ via the trilinear attention mechanism \cite{OCN}:

\begin{equation}
    f_{O_{k, l}}^Q = Attn(f_{O_k}^Q, f_{O_l}^Q) f_{O_l}^Q
\end{equation}

\begin{equation} \label{eq:attention}
     Attn(f_{O_k}^Q, f_{O_l}^Q) = \left[ \frac{exp(s_{ij})}{\sum_{j} exp(s_{ij})} \right]_{ij} \in \mathbb{R}^{n_{Qk} \times n_{O_k}}
\end{equation}

\begin{equation}
    s_{ij} = W_A^T [f_i^{O_k}; f_j^{O_l}; f_i^{O_k}\odot f_j^{O_l}] \in \mathbb{R}
\end{equation}
where $W_A \in \mathbb{R}^{3d}$ is the learnable parameters. In order to know which option is correct and which options should be eliminated, we define the keep-eliminate mechanism as:

\begin{equation}
    \tilde{f}^Q_{O_{k, l}} = [f^Q_{O_k} - f_{O_{k, l}}^Q; f^Q_{O_k} \odot f_{O_{k, l}}^Q] \in \mathbb{R}^{n_{Qk} \times 2d}
\end{equation}
The first component $(f^Q_{O_k} - f_{O_{k, l}}^Q)$ approaches vector $0$ in every dimension if option $O_k$ and $O_l$ share the same thing, while the second component $(f^Q_{O_k} \odot f_{O_{k, l}}^Q)$ is scaled according to how these two options are similar or distinguished. Then the feature for choosing or eliminating $O_k$ given the others is determined as:

\begin{equation}
    \bar{f}^Q_{O_{k}} = tanh(W_C^T [f^Q_{O_k}; \{\tilde{f}^Q_{O_{k, l}}\}_{l \ne k}] + b_C) \in \mathbb{R}^{n_{Qk} \times d}
\end{equation}
where $W_C \in \mathbb{R}^{4d \times d}$ and $b_C \in \mathbb{R}^{d}$.

However, after various transformation described above, the original information of $f^Q_{O_k}$ might be lost. To retain this information, we conduct the gating mechanism on the final representation of option $O_k$ when being compared to option $O_l$:

\begin{equation}
    f^O_k = g_k \odot f_{O_k}^Q + (1 - g_k) \odot \bar{f}^Q_{O_{k}}
\end{equation}

\begin{equation}
    g_k = sigmoid(W_g[f_{O_k}^Q, \bar{f}^Q_{O_k}] + b_g) \in \mathbb{R}^{n_{Qk}}
\end{equation}

In order to integrate the information from the context $P$ for the final decision, the cross attention is applied on every feature vector $f_k^O$ and $f_P$ to have:

\begin{equation}
    A^O_k = Attn(f_k^O, f_P) \in \mathbb{R}^{n_{Qk} \times n_P}
\end{equation}

\begin{equation}
    A^P_k = Attn(f_P, f_k^O) \in \mathbb{R}^{n_P \times n_{Qk}}
\end{equation}
These two attention matrices are used to determined $\hat{f}^O_k$ which collect the information from the context to highlight the important words in the option $O_k$:

\begin{equation}
    \hat{f}^O_k = A^O_k [f_P; A^P_k f^O_k] \in \mathbb{R}^{n_{Qk} \times 2d}
\end{equation}

Finally, the self-attention mechanism is applied to determine the final representation for the option $O_k$ according to the information of the context and keep-eliminate mechanism as follows:

\begin{equation}
    f_k = ReLU(W_O^T[\tilde{f}_k; \bar{f}_k; \tilde{f}_k - \bar{f}_k; \tilde{f}_k \odot \bar{f}_k;] + b_O)
\end{equation}
where
\begin{equation}
    \bar{f}_k = Attn(\tilde{f}_k, \tilde{f}_k) \tilde{f}_k \in \mathbb{R}^{n_{Qk} \times d}
\end{equation}
and
\begin{equation}
    \tilde{f}_k = ReLU(W_P^T[f_k^O; \hat{f}^O_k] + b_P) \in \mathbb{R}^{n_{Qk} \times d}
\end{equation}
with $W_O \in \mathbb{R}^{4d \times d}, W_P \in \mathbb{R}^{2d \times d}$ are the learnable parameters and $b_O, b_P \in \mathbb{R}^{d}$ are the bias vector.

\subsection{Option Head and Explanation Generator}

After obtained the fine-grained vectors representing four options $O_k$ $(k \in \{1, 2, 3, 4\})$, we design the Option Head module to decide which option is the correct one. In particular, the probability of \textbf{option $O_k$ is the correct option} is calculated as:

\begin{equation}
    P(k | Q, \{O_k\}_{k \in \{1, 2, 3, 4\} }, P) = \frac{exp(s_k)}{\sum_{i=1}^4 exp(s_i)}
\end{equation}
where
$$
    s_k = W^T_s MaxPool(f_k) \in \mathbb{R}
$$
with $W_s \in \mathbb{R}^{d}$ is the learnable parameters. The final option is one corresponding to the maximum $s_k$. 

Let $f_s^{max}$ be the feature vector representing the final option, then the explanation $E = \{ w^E_1, w^E_2, ..., w^E_{n_E} \}$ is generated conditioned on $f_{enc}$ as:

\begin{equation}
    P(E | f_{enc}) = \prod_{i=1}^{n_E} P(w^E_i | w^E_{<i}, f_{enc})
\end{equation}
where
\begin{equation}
    f_{enc} = [f_Q, f_P, f_k]
\end{equation}

\subsection{Objective Function}

The proposed method ViMultiChoice is trained learn multitasking which is multiple choice and explanation. To have this multitask training purpose, we define the multitask objective function as follows:

\begin{equation}
    L = L_{MC}(\theta) + L_E(\theta)
\end{equation}
where
\begin{equation}
    L_{MC}(\theta) = -\frac{1}{N} \sum_{i=1}^N ln P_\theta(k^*_i | Q_i, O_i, P_i)
\end{equation}
is the cross-entropy loss for multiple choice and

\begin{equation}
    L_E(\theta) = -\frac{1}{N} \sum_{i=1}^N \sum_{t=1}^T log P_\theta(w_{i, t}^{E*} | w_{i, <t}^{E*}, f_i^{enc})
\end{equation}

is the cross-entropy loss for explanation generation with $k_i^*$ and $w_i^{E*} = \{ w_{i, 1}^{E*}, w_{i, 2}^{E*}, ..., w_{i, T}^{E*} \}$ is is the correct option and the ground truth of explanation of the question $i^{th}$ $i^{th}$, respectively. In the following Sections, we show that multitask training enhance significantly the results of multiple choice task.

\section{Experiments}

\subsection{Datasets and Metrics}

\subsubsection{Datasets} We evaluated our proposed method against other baselines in two MCRC corpora in Vietnamese: ViMMRC 2.0 \cite{vimmrc} and our ViRCSoSciD dataset. The ViMMRC 2.0 dataset contains 5,273 multiple-choice questions in literature for students from primary school to high school in Vietnam. On the ViMMRC dataset we evaluated the reading comprehension by determining the correct option while on the ViRCSoSciD we evaluated both reading comprehension and explanation ability.

\subsubsection{Metrics} We followed previous studies \cite{vimmrc} to have F1-macro and Accuracy for evaluating the baselines and our proposed method in multiple-choice option decision. For explanation generation task, we followed studies in text summarization to deploy BLEU-4 \cite{bleu} and ROUGE-L \cite{rouge} to measure how much the explanation given by baselines and our ViMultiChoice is closed to the ground truth explanation.

\subsection{Baselines}

On our ViRCSoSciD dataset, we evaluated two classes of baselines: Large Language Models (LLMs) and the standard neural-based methods for MCRC. In particular, we evaluated open-source LLMs which have ability of reading and understanding Vietnamese: Gemini-Flash \cite{germini}, Mistral-large \cite{mistral}, QWen2 \cite{qwen2}, Vicuna \cite{vicuna}, URA-LlaMA \cite{ura-llama}, Vina-LlaMA \cite{vinallama}, and Flan-T5 \cite{flan-t5}. These LLMs are zero-shot prompted on the ViRCSoSciD dataset to perform two tasks: multiple-choice reading comprehension and explanation.

For standard neural-based methods, we evaluated three groups of baselines: 

\begin{itemize}
    \item \textbf{Bidirectional Semantic Matching:} DCMN+ \cite{dcmn}, CNN for MCRC \cite{cnn}, HAF \cite{HierarchicalAttention}, MMN \cite{mmm}, and DynSAN \cite{dynsan}.
    \item \textbf{Three-Directional Matching:} Co-Match \cite{co-matching} and CTM \cite{ctm}.
    \item \textbf{Option Comparison and Elimination:} OCN \cite{OCN} and ElimiNet \cite{ElimiNet}.
\end{itemize}

\subsection{Configuration}

All baseline models and the proposed method ViMultiChoice were trained using the Adam optimizer \cite{adam}. For baseline models based on static word embeddings, we employed pretrained word embedding PhoW2V \cite{phow2v} at syllable level, while the remaining baselines utilized the pretrained language model ViT5 \cite{vit5} to encode input sequences. Due to the maximum input length constraint of mBERT (512 tokens), the lengths of the passage, question, and answer options were capped at 400, 80, and 20 tokens, respectively, to preserve essential information for the MCRC task. 
All methods were trained with a learning rate of $5 \times 10^{-5}$, a batch size of 32 on an NVIDIA A100 GPU. The interruption of the training process is evolve if there is no improvement in F1-macro after 10 consecutive epochs. 

Moreover, we deployed an information retriever to obtain the discrete sentences for training baselines and ViMultiChoice method on the ViRCSoSciD dataset. Through our experiments, we found that VietBGE, the pretrained BGE-M3 \cite{m3-bge} specially fine-tuned for Vietnamese is the most effective retriever.

\section{Results}

\subsection{Results of LLMs on the ViRCSoSciD Dataset}

\begin{table}[htp]
    \centering
    \begin{tabular}{clcc}
    \hline
    \textbf{\#} & \textbf{Model} & \textbf{Accuracy} & \textbf{F1-macro} \\ \hline
    2 & Flan-T5 & 25.25 & 19.84 \\ 
    3 & Vicuna & 32.58 & 22.11 \\
    4 & Ura-Llama & 31.72 & 23.59 \\
    5 & Vina-Llama & 38.81 & 29.98 \\
    6 & Qwen2 & 53.94 & 13.61 \\
    7 & Gemini-Flash & \textbf{71.39} & 57.26 \\
    8 & Mistral-Large & 64.49 & \textbf{64.47} \\ \hline
    \end{tabular}
    \caption{Results (\%) of LLMs on the multiple-choice task of the ViRCSoSciD dataset. \textbf{Bold numbers} indicate the best results.}
    \label{tab:llm-mc}
\end{table}

Table \ref{tab:llm-mc} denotes the results of LLMs for the multiple choice task while Table \ref{tab:llm-explanation} indicates the results of LLMs for explanation task on our ViRCSoSciD dataset. Mitral-Large obtained the best performance on F1-macro on multiple-choice task and BLEU-4 as well as ROUGE-L on explanation generator. Gemini-Flash achieved the second position on both tasks, especially it has the best accuracy score on giving options for multiple-choice.

\begin{table}[htp]
    \centering
    \begin{tabular}{clcc}
    \hline
    \textbf{\#} & \textbf{Model} & \textbf{BLEU-4} & \textbf{ROUGE-L} \\ \hline
    2 & Flan-T5 & 0.27 & 8.66 \\ 
    3 & Vicuna & 11.83 & 34.63 \\
    4 & Ura-Llama & 12.46 & 35.93 \\
    5 & Vina-Llama & 13.83 & 36.75 \\
    6 & Qwen2 & 12.08 & 33.00 \\
    7 & Gemini-Flash & 13.60 & 41.36 \\
    8 & Mistral-Large & \textbf{16.56} & \textbf{42.23} \\ \hline
    \end{tabular}
    \caption{Results (\%) of LLMs on the explanation generation task of the ViRCSoSciD dataset. \textbf{Bold numbers} indicate the best results.}
    \label{tab:llm-explanation}
\end{table}

However, the performance of other LLMs is left far away from the two best LLMs although they were pretrained on extremely huge corpora. Moreover, Mistral-Large still gives a slightly low ROUGE-L score for explanation task. These numbers highlights the challenging of our dataset as well as the need of developing more effective LLM particularly for Vietnamese and other low-resource languages.

\begin{table}[htp]
    \centering
    \resizebox{\textwidth}{!}{
    \begin{tabular}{lcccc}
    \hline
    \textbf{Model} &
    \textbf{Literature} &
    \textbf{History} &
    \textbf{Geography} &
    \textbf{Civic Education} \\ \hline
    Flan-T5       & 25.27 & 20.31 & 17.33 & 19.97 \\
    Vicuna        & 18.73 & 24.50 & 15.16 & 25.96 \\
    Ura-Llama     & 26.19 & 28.04 & 19.78 & 28.00 \\
    Vina-Llama    & 26.29 & 39.51 & 29.26 & 35.38 \\
    Qwen2         & 18.18 & 18.99 & 22.81 & 31.44 \\
    Gemini-Flash  & \textbf{66.22} & 57.07 & 54.83 & 60.85 \\
    Mistral-Large & 52.41 & \textbf{60.04} & \textbf{63.40} & \textbf{72.59} \\
    \hline
    \end{tabular}}
    \caption{F1-macro (\%) of LLMs on the multiple-choice task of the ViRCSoSciD dataset across different subjects. \textbf{Bold numbers} indicate the best results.}
    \label{tab:llm-halfwidth}
\end{table}

Across subjects in the multiple-choice setting (Table \ref{tab:llm-halfwidth}), we observe a clear performance gap between earlier open-source models and more recent large-scale or proprietary models. Smaller instruction-tuned models such as Flan-T5 and Vicuna consistently achieve low accuracy and F1-macro scores, particularly in Geography and Civic Education, indicating limited capability in Vietnamese social science reasoning. Mid-sized open-source models (Ura-Llama, Vina-Llama, and Qwen2) show moderate improvements, with Vina-Llama generally outperforming others in this group. However, their performance remains substantially lower than that of Gemini-Flash and Mistral-Large, suggesting that model scale and training data diversity play a crucial role in multiple-choice machine reading comprehension for domain-specific Vietnamese datasets.

Among all evaluated models, Gemini-Flash and Mistral-Large consistently deliver the strongest results across all subjects, though with different strengths. Gemini-Flash achieves the highest accuracy and F1-macro in Literature, History, Geography, and Civic Education, demonstrating strong factual understanding and option selection ability. In contrast, Mistral-Large shows particularly competitive F1-macro scores, especially in History, Geography, and Civic Education, reflecting more balanced class-level predictions. The relatively smaller gap between accuracy and F1-macro for Mistral-Large suggests more stable performance across answer choices, whereas Gemini-Flash appears to benefit from stronger confidence in dominant classes.

\begin{table}[htp]
    \centering
    \resizebox{\textwidth}{!}{
    \begin{tabular}{lcccccccc}
    \hline
    \textbf{Model} &
    \multicolumn{2}{c}{\textbf{Literature}} &
    \multicolumn{2}{c}{\textbf{History}} &
    \multicolumn{2}{c}{\textbf{Geography}} &
    \multicolumn{2}{c}{\textbf{Civic Education}} \\
     & \textbf{B-4} & \textbf{R-L} & \textbf{B-4} & \textbf{R-L} & \textbf{B-4} & \textbf{R-L} & \textbf{B-4} & \textbf{R-L} \\
    \hline
    Flan-T5       & 0.25  & 8.64  & 0.18  & 6.64  & 0.40  & 11.14 & 0.24  & 8.17 \\
    Vicuna        & 9.11  & 32.24 & 9.26  & 32.61 & 9.65  & 30.89 & 17.71 & 41.59 \\
    Ura-Llama     & 11.88 & 36.34 & 7.85  & 32.10 & 11.09 & 34.35 & 19.35 & 41.96 \\
    Vina-Llama    & 11.25 & 34.88 & 12.19 & 36.70 & 12.07 & 34.32 & 18.19 & 39.89 \\
    Qwen2         & 9.94  & 31.77 & 11.86 & 35.71 & 11.40 & 31.74 & 13.51 & 31.61 \\
    Gemini-Flash  & 8.79  & \textbf{42.17} & 10.57 & 38.74 & 14.34 & 41.34 & 17.16 & 44.18 \\
    Mistral-Large & \textbf{13.35} & 41.97 & \textbf{13.04} & \textbf{39.54} & \textbf{16.04} & \textbf{41.60} & \textbf{21.76} & \textbf{46.03} \\
    \hline
    \end{tabular}}
    \caption{BLEU-4 (B-4) (\%) and ROUGE-L (R-L) (\%) of LLMs on the explanation generation task of the ViRCSoSciD dataset across different subjects. \textbf{Bold numbers} indicate the best results.}
    \label{tab:llm-generation-halfwidth}
\end{table}

For the explanation generation task (Table \ref{tab:llm-generation-halfwidth}), a similar performance hierarchy is observed, but with more nuanced differences. Mistral-Large consistently achieves the best BLEU-4 and ROUGE-L scores across all subjects, indicating superior ability in generating coherent and content-aligned explanations. While Gemini-Flash performs competitively—especially in ROUGE-L—it generally lags behind Mistral-Large in BLEU-4, suggesting less precise lexical overlap despite producing fluent explanations. In contrast, earlier models such as Flan-T5 yield extremely low scores, highlighting their difficulty in generating meaningful explanatory text in Vietnamese. Overall, these results indicate that strong multiple-choice accuracy does not necessarily translate into high-quality explanation generation, underscoring the importance of advanced generative capabilities for explanation-based reasoning tasks.

\subsection{Results of ViMultiChoice}

\subsubsection{Results of Retrieval Methods}

\begin{table}[htp]
    \centering
    \renewcommand{\arraystretch}{1.5} 
    \resizebox{\textwidth}{!}{
    \begin{tabular}{l|ll|ll|ll|ll}
    \hline
    \textbf{Retriever} & \multicolumn{1}{c}{\textbf{P@10}} & \multicolumn{1}{c|}{\textbf{R@10}} & \multicolumn{1}{c}{\textbf{P@15}} & \multicolumn{1}{c|}{\textbf{R@15}} & \multicolumn{1}{c}{\textbf{P@20}} & \multicolumn{1}{c|}{\textbf{R@20}} & \multicolumn{1}{c}{\textbf{P@30}} & \multicolumn{1}{c}{\textbf{R@30}} \\ \hline
    BM25 & 28.04 & 03.75 & 24.29 & 04.77 & 21.89 & 05.66 & 18.68 & 07.11 \\
    MiniLML12 & 29.42 & 03.92 & 26.57 & 05.23 & 24.61 & 06.39 & 21.85 & 08.39 \\
    MultiE5 & 38.52 & 05.25 & 34.07 & 06.86 & 31.05 & 08.24 & 26.94 & 10.56 \\
    VietBGE & \textbf{42.77} & \textbf{05.78} & \textbf{38.12} & 07.60 & \textbf{34.75} & \textbf{09.12} & \textbf{30.16} & {\ul 11.69} \\
    VietBGEMultiE5 & {\ul 42.53} & {\ul 05.75} & {\ul 37.92} & {\ul 07.57} & {\ul 34.56} & {\ul 09.10} & {\ul 30.12} & \textbf{11.71} \\ \hline
    \end{tabular}}
    \caption{Results (\%) of Retrievers. \textbf{Bold numbers} indicate the best results while \underline{underlined numbers} indicate results within 0.1\% from the best.}
    \label{tab:retriever}
\end{table}

We employed \textbf{Precision@K (P@K)} and \textbf{Recall@K (R@K)} to evaluate the relevance between the retrieved context and the target lesson content associated with each question. This evaluation is conducted independently for both sparse and dense retrieval methods. Subsequently, we apply the \textbf{Reciprocal Rank Fusion (RRF)} algorithm to integrate the results from both approaches, investigating whether this hybrid integration enhances overall retrieval performance. 

The results presented in Table~\ref{tab:retriever} indicate that integrating the two optimal retrieval sources does not yield any performance improvements. Conversely, this hybrid approach significantly increases computational overhead, highlighting an inefficient trade-off between resource consumption and retrieval quality. Furthermore, we set $K = 15$ for context construction to strike an optimal balance between enhancing contextual diversity and maintaining retrieval performance.

\subsubsection{Results on the MCRC tasks}

\begin{table}[htp]
    \centering
    \begin{tabular}{lcccc}
    \hline
    \multirow{2}{*}{\textbf{Model}} & \multicolumn{2}{c}{\textbf{ViRCSoSciD}} & \multicolumn{2}{c}{\textbf{ViMMRC 2.0}} \\ 
     & \textbf{Acc} & \textbf{F1-macro} & \textbf{Acc} & \textbf{F1-macro} \\ \hline
    \multicolumn{5}{l}{\textit{Published results$^\dagger$}} \\
    mBERT & - & - & 47.79 & - \\
    mBERT + MAN & - & - & 52.93 & - \\
    mBERT + NLI & - & - & 56.99 & - \\
    mBERT + MAN + NLI & - & - & 55.64 & - \\
    XLM-R & - & - & 51.84 & - \\
    viBERT & - & - & 53.74 & - \\
    viBERT + MAN & - & - & 57.17 & - \\
    viBERT + NLI & - & - & 56.81 & - \\
    viBERT + MAN + NLI & - & - & 58.81 & - \\
    BERT4News & - & - & 41.74 & - \\ \hline
    \multicolumn{5}{l}{\textit{Our implementations}} \\
    HAF & 26.97 & 26.49 & 31.65 & 30.68 \\
    CNN for MCQA & 30.09 & 30.09 & 35.08 & 34.73 \\
    CoMatch & 34.76 & 33.86 & 41.21 & 40.55 \\
    ElimiNET & 33.13 & 32.90 & 38.59 & 38.21 \\
    MMN & 41.23 & 41.20 & 52.93 & 52.73 \\
    DynSAN & 43.65 & 43.54 & 44.36 & 43.74 \\
    CTM & 50.43 & 50.36 & 57.17 & 57.17 \\
    DCMN+ & 53.55 & 53.56 & 55.73 & 55.71 \\  \hline
    ViT5 Encoder & 59.08 & 59.07 & 56.63 & 56.54 \\
    OCN & 54.64 & 54.61 & 56.27 & 56.29 \\
    ViMultiChoice (w/o Expl.) & {\ul 62.98} & {\ul 63.03} & \textbf{60.32} & \textbf{60.23} \\
    ViMultiChoice & \textbf{64.38} & \textbf{64.38} & - & - \\ \hline
    \end{tabular}
    \caption{Results of ViMultiChoice and baselines for multiple-choice reading comprehension. \textbf{Bold numbers} indicate the best scores. $^\dagger$Results reported from the study of [\textit{Luu et al., 2023}]. NLI stands for Natural Language Inference and MAN stands for Multi-step Attention Network. \textbf{w/o Expl.} stands for ViMultiChoice without the Explanation Generator.}
    \label{tab:vimultichoice-mc}
\end{table}

On the other hands, the results of standard neural network baselines are more impressive. As indicated in Table \ref{tab:vimultichoice-mc}, OCN method achieved the highest Accuracy and F1-macro on the ViRCSoSciD dataset. Moreover, baselines such as OCN, CTM, and DCMN+ approaches the performance of LLMs, except for Gemini-Flash and Mitral-Large, on multiple-choice task of the ViRCSoSciD dataset.

Especially, in comparison to the original OCN, our ViMultiChoice can easily obtained the best Accuracy and F1-macro metrics on both ViMMRC 2.0 and ViRCSoSciD datasets. Moreover, on the ViMMRC 2.0 dataset, our proposed ViMultiChoice has significantly better scores than multi-stage baselines reported from the study \cite{vimmrc}. These results strongly highlights the effectiveness of the ViWordFormer module for modeling phrasal lexemes of Vietnamese texts, which enhance the reading and understanding ability of the original OCN for multiple-choice task.

In addition, our ViMultiChoice has improved by a large margins when being trained under the multitask objective functions. According to Table \ref{tab:vimultichoice-mc}, learning how to explain the option decision enhances the performance of ViMultiChoice from \textbf{62.98\%} to \textbf{64.38\%} on Accuracy and from \textbf{63.03\%} to \textbf{64.38\%} on F1-macro.

\section{Ablation study of ViMultiChoice}

The novelty of our ViMultiChoice is the introduction of the ViWordFormer module for improving Vietnamese text representation and the Explanation Generator for providing explanation of the decision. To highlight the effectiveness of the ViWordFormer module as well as the advantage of having the Explanation Generator, we conducted ablation studies and provided the evaluation in Table \ref{tab:ablation}.

\begin{table}[htp]
    \centering
    \resizebox{\textwidth}{!}{
    \begin{tabular}{lccccccc}
    \hline
    \multirow{2}{*}{\textbf{Dataset}} & \multirow{2}{*}{\textbf{VWF}} & \multirow{2}{*}{\textbf{Type}} & \multicolumn{2}{c}{\textbf{Multiple Choice}} & \multicolumn{2}{c}{\textbf{Explanation}} \\
     &  &  & \textbf{Accuracy} & \textbf{F1-macro} & \textbf{BLEU-4} & \textbf{ROUGE-L} \\ \hline
    \multirow{2}{*}{ViRCSoSciD} & \xmark & \multirow{4}{*}{Single Task} & 61.81 & 61.79 & - & - \\
    & \cmark &  & \textbf{62.98} & \textbf{63.03} & - & - \\ 
    \multirow{2}{*}{ViMMRC 2.0} & \xmark &  & 56.27 & 56.29 & - & - \\
    & \cmark &  & \textbf{60.32} & \textbf{60.23} & - & - \\ \hline
    \multirow{2}{*}{ViRCSoSciD} & \xmark & \multirow{2}{*}{Multitask} & 62.51 & 62.51 & 20.23 & 44.71 \\
    & \cmark &  & \textbf{64.38} & \textbf{64.38} & \textbf{34.85} & \textbf{56.78} \\ \hline
    \end{tabular}}
    \caption{Ablation study of ViWordFormer (VWF) module for ViMultiChoice on the ViRCSoSciD and ViMMRC 2.0 datasets.}
    \label{tab:ablation}
\end{table}

Results from Table \ref{tab:ablation} indicate that having the ViWordFormer module, ViMultiChoice obtained better scores on both Accuracy and F1-macro for multiple-choice task on the ViRCSoSciD and ViMMRC 2.0 datasets. Moreover, ViWordFormer module also improves the performance of ViMultiChoice on explanation generation task by a large margin, from \textbf{20.23\% to 34.85\%} on BLEU-4 metric and from \textbf{44.71\% to 56.78\%} on ROUGE-L metric.

\section{Results Analysis of ViMultiChoice}

\begin{table}[htp]
    \centering
    \resizebox{\textwidth}{!}{
    \begin{tabular}{lccccc}
    \hline
    \textbf{Model} & \textbf{WM} & \textbf{Para} & \textbf{SSR} & \textbf{MSR} & \textbf{\begin{tabular}[c]{@{}c@{}}Ambiguous/\\ Insufficient\end{tabular}} \\ \hline
    HAF & 22.50 & 26.98 & 30.25 & 32.51 & 36.13 \\
    ElimiNET & 35.00 & 38.10 & 37.72 & 39.60 & 36.97 \\
    CNN for MCQA & 42.50 & 41.27 & 34.16 & 35.81 & 27.73 \\
    MMN & 80.00 & 57.14 & 59.07 & 49.67 & 43.70 \\
    DynSAN & 75.00 & 42.86 & 50.53 & 41.75 & 33.61 \\
    DCMN+ & 92.50 & 61.90 & 62.99 & 51.65 & 43.70 \\
    CTM & 92.50 & 61.90 & 64.41 & 53.63 & 43.70 \\ \hline
    ViT5 Encoder & 87.50 & 60.32 & 63.35 & 54.29 & 40.34 \\
    OCN & 92.50 & 65.08 & 64.77 & 51.65 & 42.86 \\ 
    ViMultiChoice (w/o Expl.) & \textbf{95.00} & \textbf{65.08} & \textbf{66.55} & \textbf{57.43} & \textbf{46.22} \\ \hline
    \end{tabular}}
    \caption{Accuracy of ViMultiChoice and baselines on each question type of the ViMMRC 2.0 dataset. \textbf{WM} stands for Word Matching. \textbf{Para} stands for Paraphrasing. \textbf{SSR} stands for Single-Sentence Reasoning. \textbf{MSR} stands for Multiple-Sentence Reasoning. \textbf{w/o Expl.} stands for ViMultiChoice without Explanation Generator.}
    \label{tab:results-question}
\end{table}

As shown in Table \ref{tab:results-question}, ViMultiChoice achieves the highest accuracy across all question types in the ViMMRC 2.0 dataset. Specifically, OCN attains accuracies of 64.77\% on Single-Sentence Reasoning and 51.65\% on Multiple-Sentence Reasoning questions, while its backbone model, ViT5 Encoder, achieves 63.35\% and 54.29\%, respectively. In contrast, ViMultiChoice delivers substantially better performance on both question types, with accuracies of 66.55\% for Single-Sentence Reasoning and 57.43\% for Multiple-Sentence Reasoning.

\begin{table}[htp]
    \centering
    \begin{tabular}{lccc}
    \hline
    \textbf{Model} & \textbf{Grade 10} & \textbf{Grade 11} & \textbf{Grade 12} \\ \hline
    HAF & 26.71 & 26.61 & 26.26 \\
    CNN for MCQA & 29.44 & 29.12 & 30.45 \\
    CoMatch & 32.77 & 37.56 & 32.96 \\
    ElimiNET & 30.17 & 31.15 & 33.98 \\
    MMN & 42.16 & 42.93 & 40.36 \\
    DynSAN & 42.83 & 41.31 & 44.11 \\
    CTM & 43.67 & 53.83 & 50.78 \\ 
    DCMN+ & 53.43 & 55.38 & 53.01 \\ \hline
    ViT5 Encoder & 54.13 & 63.77 & 58.85 \\
    OCN & 56.60 & 54.23 & 54.11 \\
    ViMultiChoice (w/o Expl.) & \textbf{61.85} & 67.10 & 62.14 \\
    ViMultiChoice & \underline{60.12} & \textbf{67.43} & \textbf{64.48} \\ \hline
    \end{tabular}
    \caption{F1-macro of ViMultiChoice and baselines on each grade of the ViRCSoSciD dataset. \textbf{w/o Expl.} stands for ViMultiChoice without Explanation Generator.}
    \label{tab:results-grade}
\end{table}

\begin{table}[htp]
    \centering
    \resizebox{\textwidth}{!}{
    \begin{tabular}{lcccc}
    \hline
    \textbf{Model} & \textbf{Literature} & \textbf{History} & \textbf{Geography} & \textbf{Civic Education} \\ \hline
    CNN for MCQA & 37.80 & 32.55 & 24.80 & 31.40 \\
    CoMatch & 26.88 & 37.42 & 30.91 & 34.14 \\
    ElimiNET & 28.27 & 31.07 & 32.68 & 36.02 \\
    DCMN+ & 45.08 & 50.66 & 51.21 & 60.86 \\
    MMN & 35.25 & 41.05 & 39.13 & 44.73 \\
    DynSAN & 36.80 & 42.26 & 40.25 & 49.43 \\
    CTM & 39.19 & 46.74 & 50.50 & 56.16 \\
    HAF & 22.02 & 28.44 & 25.85 & 25.63 \\
    OCN & 48.27 & 54.78 & 53.09 & 57.22 \\ \hline
    ViMultiChoice (w/o Expl.) & 58.29 & \textbf{61.28} & 59.70 & 69.88 \\
    ViMultiChoice & \textbf{66.19} & 60.35 & \textbf{62.98} & \textbf{70.18} \\ \hline
    \end{tabular}}
    \caption{F1-macro of ViMultiChoice and baselines on each grade of the ViRCSoSciD dataset. \textbf{w/o Expl.} stands for ViMultiChoice without Explanation Generator.}
    \label{tab:results-subject}
\end{table}

On the ViRCSoSciD dataset, ViMultiChoice attains the highest macro F1 score across all Vietnamese high school grade levels (Table \ref{tab:results-grade}). It substantially outperforms ViT5 Encoder and OCN on Grade~10 and Grade~12 questions, indicating the effectiveness of the proposed approach. Subject-level analysis (Table~\ref{tab:results-subject}) further shows that ViMultiChoice achieves the best macro F1 in most subjects, particularly Civic Education. With the Explanation Generator, the model improves performance across nearly all subjects, except History, where a slight decrease is observed. Overall, ViMultiChoice under both single-task and multi-task training settings demonstrates clear improvements over all baseline models.

\section{Conclusion}

In this study, we proposed ViMultiChoice, a novel method tailored to Vietnamese linguistic characteristics for multiple-choice reading comprehension with explanation generation. To evaluate the proposed approach, we introduced ViRCSoSciD, the first large-scale dataset of multiple-choice questions with human-provided explanations in Vietnamese. Through extensive experiments, we demonstrated that jointly training ViMultiChoice to generate explanations leads to improved performance in selecting the correct option.

\section*{Ethical Statement}

The ViRCSoSciD has been sourced from public resources that allow for redistribution. All instances in the ViRCSoSciD dataset have undergone a thorough review to ensure the exclusion of any examples that raise ethical concerns.

\bibliographystyle{elsarticle-num-names}
\bibliography{main}

\newpage

\appendix

\section{Detailed Configurations of LLMs}

\begin{table}[ht]
    \centering
    \begin{tabular}{clrr}
    \hline
    \# & \textbf{Model} & \textbf{Version} & \textbf{\# Parameters} \\ \hline
    1 & Flan-T5 & google/flan-t5-large & 780M \\
    2 & Vicuna & lmsys/vicuna-7b-v1.5 & 7B \\
    3 & Ura-Llama & ura-hcmut/ura-llama-7b & 7B \\
    4 & Vina-Llama & vilm/vinallama-7b-chat & 7B \\
    5 & Ghost & ghost-x/ghost-7b-alpha & 7B \\
    6 & Qwen2 & Qwen2-7B-Instruct & 7.6B \\
    7 & Gemini-Flash & gemini-1.5-flash & 8B \\
    8 & Mistral-Large & mistral-large-2 & 123B \\ \hline
    \end{tabular}
    \caption{Detailed versions of LLMs in our evaluation.}
    \label{tab:llm-version}
\end{table}

The detailed versions of LLMs in this study are provided in Table \ref{tab:llm-version}. The hyperparameter configurations are set the same for all LLMs as follows:
\begin{itemize}
    \item Temperature: 1.0
    \item Max new tokens: 300
    \item Top-K: 1
    \item Repetition penalty: 1.1
\end{itemize}

\begin{tcolorbox}[title=Prompt for Zero-shot Flan-T5, colback=blue!5!white, colframe=black!75!black, height=6cm]
\begin{verbatim}
Câu hỏi: <question>
Các lựa chọn: 
    'A': <option_A>
    'B': <option_B>
    'C': <option_C>
    'D': <option_D>
Hãy chọn đáp án đúng (A, B, C hoặc D), sau đó cho lời giải 
thích ngắn gọn.
Đáp án:
\end{verbatim}
\end{tcolorbox}

\begin{tcolorbox}[title=Prompt for Zero-shot Vicuna, colback=blue!5!white, colframe=black!75!black, height=8.5cm]
\begin{verbatim}
Bạn là một trợ lý thông minh. Hãy trả lời người dùng một 
cách chính xác bằng cách chọn đáp án đúng, ghi ngắn gọn 
A, B, C hay D, sau đó cho lời giải thích ngắn gọn. 

USER: 
    Câu hỏi: {question}
    Các lựa chọn: 
        'A': {option_A}
        'B': {option_B}
        'C': {option_C}
        'D': {option_D}

ASSISTANT:
    Trả lời: 
\end{verbatim}
\end{tcolorbox}

\begin{tcolorbox}[title=Prompt for Zero-shot Ura-Llama, colback=blue!5!white, colframe=black!75!black, height=8.5cm]
\begin{verbatim}
Bạn là một trợ lý thông minh. Hãy trả lời người dùng một 
cách chính xác bằng cách chọn đáp án đúng, ghi ngắn gọn 
A, B, C hay D, sau đó cho lời giải thích ngắn gọn.

USER: 
    Câu hỏi: {question}
    Các lựa chọn: 
        'A': {option_A}
        'B': {option_B}
        'C': {option_C}
        'D': {option_D}

ASSISTANT:
    Trả lời: 
\end{verbatim}
\end{tcolorbox}

\begin{tcolorbox}[title=Prompt for Zero-shot Vina-Llama, colback=blue!5!white, colframe=black!75!black, height=9.5cm]
\begin{verbatim}
system

Bạn là một trợ lí AI hữu ích. Hãy trả  lời người dùng một 
cách chính xác bằng cách chọn đáp án đúng, ghi ngắn gọn A, B, C 
hay D, sau đó cho lời giải thích ngắn gọn."

user
    Câu hỏi: {question}
    Các lựa chọn: 
        'A': {option_A}
        'B': {option_B}
        'C': {option_C}
        'D': {option_D}

assistant
Trả lời:
\end{verbatim}
\end{tcolorbox}

\begin{tcolorbox}[title=Prompt for Zero-shot Qwen2, colback=blue!5!white, colframe=black!75!black, height=7.5cm]
\begin{verbatim}
{
"role": "user", 
"content":
    Câu hỏi: {question}
    Các lựa chọn: 
        'A': {option_A}
        'B': {option_B}
        'C': {option_C}
        'D': {option_D}
    Hãy chọn đáp án đúng, ghi ngắn gọn A, B, C hay D, sau 
    đó cho lời giải thích ngắn gọn.
}
\end{verbatim}
\end{tcolorbox}

\begin{tcolorbox}[title=Prompt for Zero-shot Gemini-Flash, colback=blue!5!white, colframe=black!75!black, height=6.5cm]
\begin{verbatim}
Câu hỏi: {question}

Các lựa chọn: 
    'A': {option_A}
    'B': {option_B}
    'C': {option_C}
    'D': {option_D}

Hãy chọn đáp án đúng, ghi ngắn gọn A, B, C hay D, sau đó 
cho lời giải thích ngắn gọn.
\end{verbatim}
\end{tcolorbox}

\begin{tcolorbox}[title=Prompt for Zero-shot Mistral-Large, colback=blue!5!white, colframe=black!75!black, height=8.5cm]
\begin{verbatim}
{
"role": "user", 
"content": 

    User:
    Câu hỏi: {question}
    Các lựa chọn: 
        'A': {option_A}
        'B': {option_B}
        'C': {option_C}
        'D': {option_D}
    Hãy chọn đáp án đúng, ghi ngắn gọn A, B, C hay D, sau 
    đó cho lời giải thích ngắn gọn.
}
\end{verbatim}
\end{tcolorbox}

\section{Visualization of Results }

\begin{figure}[htp]
    \centering
    \begin{subfigure}{0.85\linewidth}
        \centering
        \includegraphics[width=\linewidth]{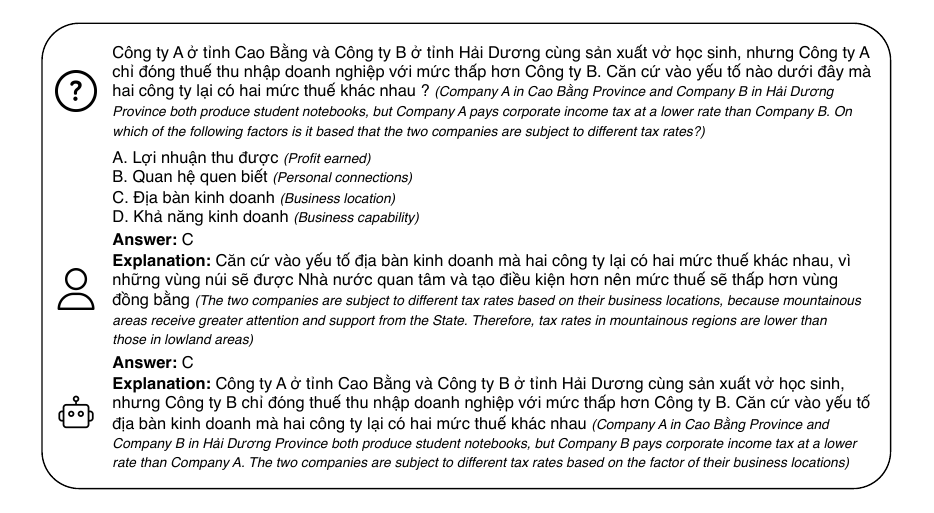}
    \end{subfigure}

    \begin{subfigure}{0.85\linewidth}
        \centering
        \includegraphics[width=\linewidth]{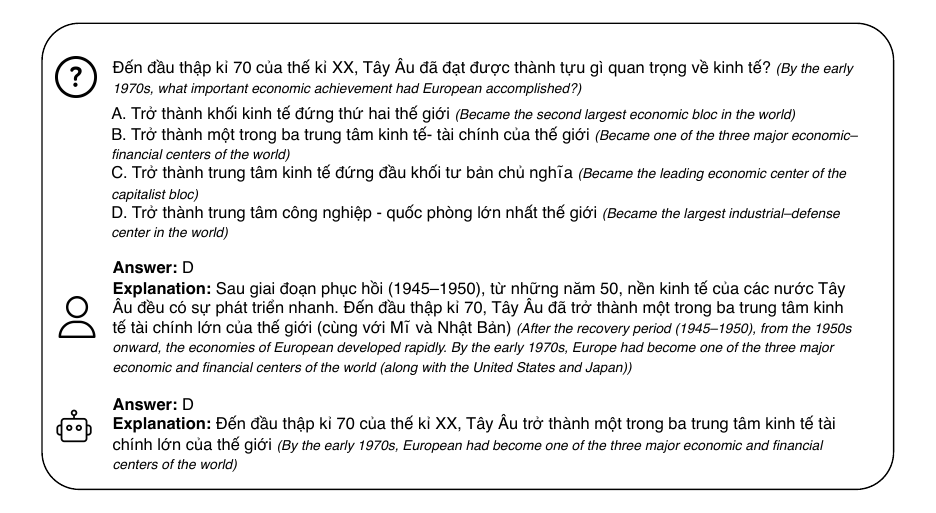}
    \end{subfigure}

    \begin{subfigure}{0.85\linewidth}
        \centering
        \includegraphics[width=\linewidth]{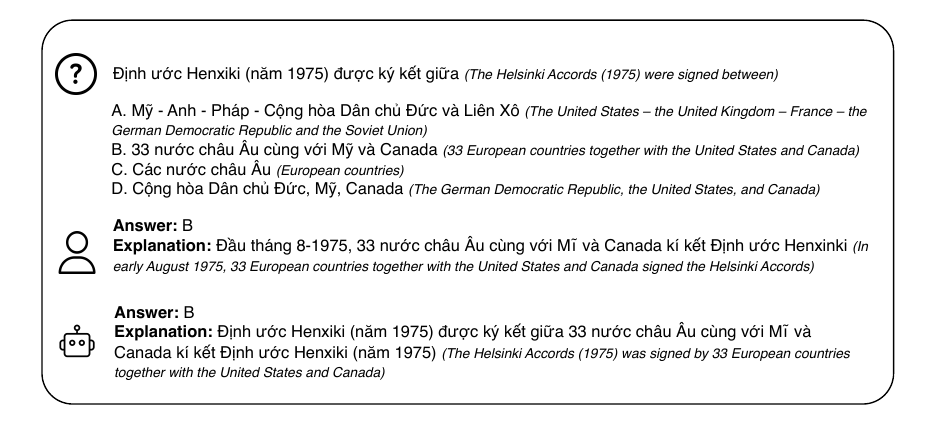}
    \end{subfigure}
    
    \caption{Examples for the correct option decision of ViMultiChoice.}
    \label{fig:true-samples}
\end{figure}

\begin{figure}[htp]
    \centering
    \begin{subfigure}{0.85\linewidth}
        \centering
        \includegraphics[width=\linewidth]{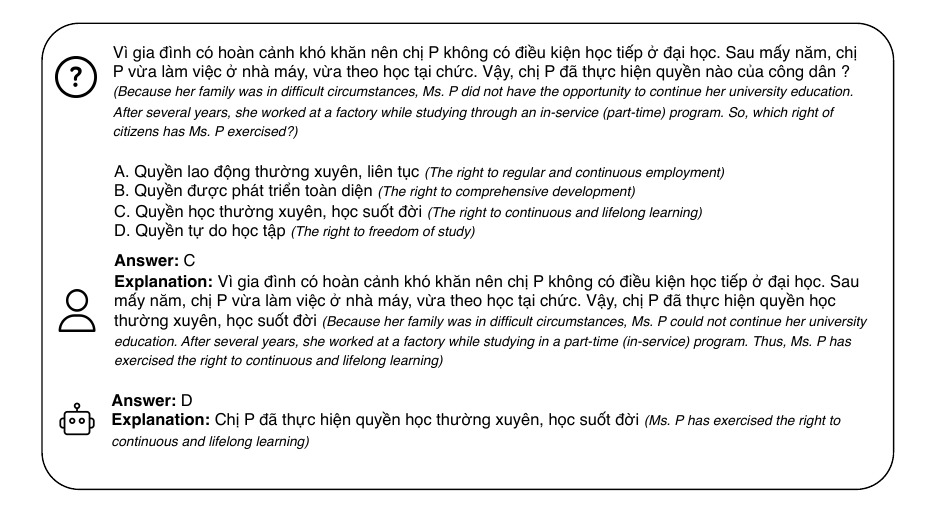}
    \end{subfigure}

    \begin{subfigure}{0.85\linewidth}
        \centering
        \includegraphics[width=\linewidth]{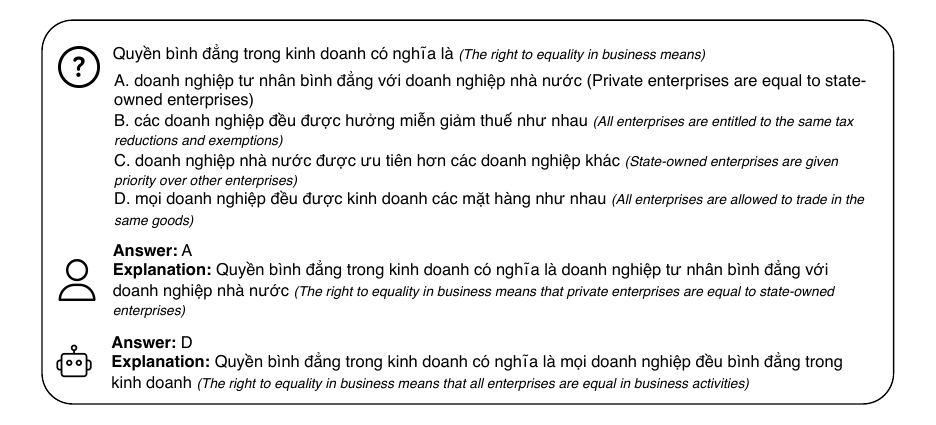}
    \end{subfigure}

    \begin{subfigure}{0.85\linewidth}
        \centering
        \includegraphics[width=\linewidth]{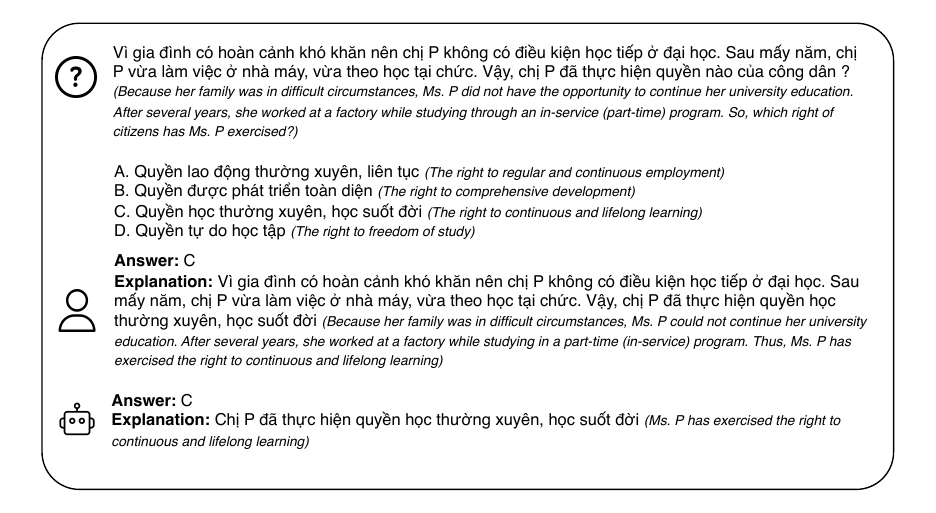}
    \end{subfigure}
    
    \caption{Examples for the wrong option decision of the ViMultiChoice.}
    \label{fig:wrong-samples}
\end{figure}

We provide some results of ViMultiChoice for qualitative demonstration. In particular, ViMultiChoice and determine correctly the topic and subject of the given question (Figure \ref{fig:true-samples}). This is indicated in the explanation it generated. This advantage help this model give the correct answers and the comprehensive explanation for its decision. However, on questions having confusing or closely similar information (Figure \ref{fig:wrong-samples}), ViMultiChoice tends to have incorrect decision even though it gave reasonable explanations. The further improvement on ViMultiChoice should rely on improving the comparison of options against each other to enhance the ability of determining correctly answer among confusing ones.

\end{document}